# Pattern Recognition

## Presentation Attack Detection Methods based on Gaze Tracking and Pupil Dynamic: A Comprehensive Survey
### --Manuscript Draft--

| | |
|---|---|
| **Manuscript Number:** | |
| **Article Type:** | Survey/Review paper |
| **Section/Category:** | Face and biometrics |
| **Keywords:** | Gaze Tracking, Pupil dynamic, Presentation Attack Detection |
| **Corresponding Author:** | Jalil Nourmohammadi Khiarak<br><br>Warsaw, POLAND |
| **First Author:** | Jalil Nourmohammadi Khiarak |
| **Order of Authors:** | Jalil Nourmohammadi Khiarak |
| **Abstract:** | Purpose of the research : In the biometric community, visible human characteristics are popular and viable for verification and identification on mobile devices. However, imposters are able to spoof such characteristics by creating fake and artificial biometrics to fool the system. Visible biometric systems have suffered a high-security risk of presentation attack.<br>Methods : In the meantime, challenge-based methods, in particular gaze tracking and pupil dynamic appear to be more secure methods than others for contactless biometric systems. We review the existing work that explores gaze tracking and pupil dynamic liveness detection.<br>The principal results : This research analyzes various aspects of gaze tracking and pupil dynamic presentation attacks, such as state-of-the-art liveness detection algorithms, various kinds of artifacts, the accessibility of public databases, and a summary of standardization in this area. In addition, we discuss future work and the open challenges to creating a secure liveness detection based on challenge-based systems. |

Cover Letter

# Presentation Attack Detection Methods based on Gaze Tracking and Pupil Dynamic: A Comprehensive Survey

**Jalil Nourmohammadi Khiarak**

Institute of Control and Computation Engineering, Warsaw University of Technology, Warsaw, Poland

Corresponding author: Jalil Nourmohammadi Khiarak (e-mail: Jalil.Nourmohammadi@elka.pw.edu.pl).

"This work was supported by the EU Horizon 2020 Framework for Research and Innovation under Grant Agreement Number 675087 (AMBER)."

Highlights (for review)

# Presentation Attack Detection Methods based on Gaze Tracking and Pupil Dynamic: A Comprehensive Survey

**Jalil Nourmohammadi Khiarak**

Institute of Control and Computation Engineering, Warsaw University of Technology, Warsaw, Poland
(e-mail: Jalil.Nourmohammadi@elka.pw.edu.pl).

**ABSTRACT:**

**Purpose of the research**: In the biometric community, visible human characteristics are popular and viable for verification and identification on mobile devices. However, imposters are able to spoof such characteristics by creating fake and artificial biometrics to fool the system. Visible biometric systems have suffered a high-security risk of presentation attack.

**Methods**: In the meantime, challenge-based methods, in particular gaze tracking and pupil dynamic appear to be more secure methods than others for contactless biometric systems. We review the existing work that explores gaze tracking and pupil dynamic liveness detection.

**The principal results**: This research analyzes various aspects of gaze tracking and pupil dynamic presentation attacks, such as state-of-the-art liveness detection algorithms, various kinds of artifacts, the accessibility of public databases, and a summary of standardization in this area. In addition, we discuss future work and the open challenges to creating a secure liveness detection based on challenge-based systems.

**INDEX TERMS** Gaze Tracking, Pupil dynamic, Presentation Attack Detection,

## 1. Introduction

Biometric verification and identification contribute a high level of safety by authenticating users based on behavioral characteristics including gait, keystroke, signature, and physiological characteristics such as iris, face, ear, and fingerprint. Biometric characteristics remove the need for the most traditional approaches of verification and identification involving PIN and password. Visible biometrics, more specifically face, iris, and ocular characteristics, are utilized in both offline and online applications. The state-of-the-art for these algorithms achieves the highest accuracy even for mobile devices. As a result, iris and facial recognition are utilized in a range of mobile devices, forensics applications, e-commerce contexts and surveillance systems. Over the last two decades, face and iris recognition have become a standard application. In addition, visible and contactless biometrics benefit low-cost sensors and nonintrusive data capture.

Given the widespread use and application of iris and facial recognition in applications, it is perhaps inevitable that there are attempts by imposters and attackers to spoof or unlock mobile devices. This is especially true if there is no Presentation Attack Detection (PAD) method, though it is likely that even if mobile device biometric systems have PAD, attackers have found ways to spoof mobile devices and unlock systems. It is well-known that Face ID, despite being a cutting-edge facial recognition system, can be tricked by masks, children, and twins. Meanwhile, German hackers managed to unlock the Samsung Galaxy S8 iris scanner in less than a month after it became publicly available. These cases indicate the vulnerability of visible and untouchable biometric recognition systems in mobile devices. Visible biometrics are available to attackers and can motivate them to spoof the systems in order to steal personal information, emails, payment information and other nefarious goals.

One well-known approach to iris liveness detection is pupil dynamics[1, 2], which is very difficult to spoof. This demonstrates that gaze tracking is well-known for face liveness detection [3, 4]. There are two different setups for gaze tracking. In the first setup – known as screen-based gaze tracking – the objective is to find out where the person is looking at the screen [5]. The second kind of gaze tracking is known as wearable gaze tracking [6]. This setup has a rather complicated system whereby the person wears glasses with a scene camera located at the center. Both systems use an image or a video frame, where the objective is to find x and y location of the person's gaze on this frame. The two setups give two different input images, one uses an eye image the other one uses the image on the screen. In gaze tracking, it is important to select direct evidence, also known as ground truths. With this in mind, there are two different methods for finding a person's gaze. The first of these is the active method, where the person is asked to fix their gaze on a particular location of the screen, often at a specific point on the image. In this way, we are able to establish a ground truth. This allows us to know the shape of the eye when it is looking at that location, so by asking the person to stare at different points in turn, we can build up ground truths about the eye. In this research, we use active model to make evaluations. In passive method, on the other hand, the user is simply asked to use their computer normally and, while using it there are certain things that require them to look at certain locations, which can then be noted and collected.

In this paper, we review iris and face PAD methods based on gaze tracking and pupil dynamics. Although there are excellent surveys for face and iris liveness detection methods, including biometric anti-spoofing handbooks, there is no survey for PAD-based eye tracking and pupil dynamics used for face and iris. Therefore, we offer an overview of recent studies on GTPD for face and iris PAD, databases, an analysis of various state-of-the-art PAD methods, standards of PAD, and evaluation metrics.

The remainder of the article is organized as follows: Section 2 introduces the human eye and challenges to pupil detection; Section 3 introduces GTPD; Section 4 discusses the literature review for GTPD and gives an overview of PAD for the face and iris using eye movement; Section 5 discuses all available databases; Section 6 discusses performance metrics according to ISO/IEC; Section 7 provides future work and the open challenges to GTPD; and the conclusions will be described in the last section.

## 2. The human eye and challenges to GTPD

The human eye has a lens that directs the lights onto a retina; it also has the iris, which controls an opening called the pupil [7]. Therefore, light goes through the pupil, reflects through the lens and an image is projected onto the retina. Figure 1 shows the details.

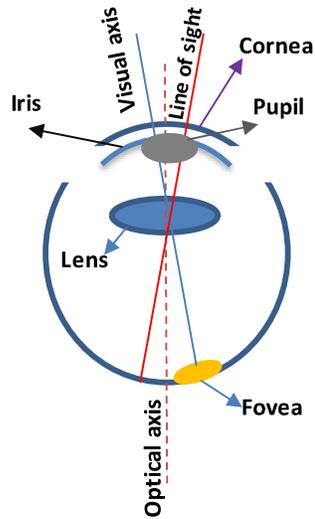

**Fig. 1**. Features of eye and definitions of the specific parts of the eye that are considered in our paper

Now the line connecting the center of the pupil to the center of the lens is called the optical axis of the eye. You might think that this is the direction of the gaze, but in fact it is not. There is a place in the retina where the image is focused, and this is where the sharpest image is formed, known as the fovea. The line connecting the fovea with the center of the lens is what determines visual gaze. We need a special calibration to estimate the visual axis given the optical axis. We should ask the person to focus on one point at a time and assume that the gaze point is there, or close to that point. Then we can calculate the offset between the real gaze and estimated gaze and store the extracted information. When next estimating the person's gaze, it is possible to fix this estimate by accounting for the person's specific calibration. When the light hits the cornea, it forms an image. The cornea reflection is appeared on some part of the iris. The line connecting the pupil to the cornea is very informative in estimating the gaze direction. An algorithm based on this approach has been proposed, called the pupil center corneal reflection. It uses an arrow from the pupil to the cornea reflection to estimate the gaze.

We can use other features of eye to gaze estimation such as iris, pupil, corneal reflection and eye counters[8]. Using the image of the eye or extracted features of the eye to estimate the direction of gaze is called gaze mapping. There is a method for gaze mapping called the 2D regression-based method, which maps 2D features without considering geometric aspects [9].

Pupil location on the eyes is the most valuable information that needs to be established. There are two different setups for pupils, namely dark pupil and white pupil. In a dark pupil setup [10], the light source and the camera are not co-located. They are located at two separate locations. This means that when capturing images of iris, the pupil will be dark. In a white pupil setup, the light source and the camera are co-located the light rebounds from the retina and creates a white spot. This is the exact same phenomenon that we see in red eye, which happened in systems where the flash is almost co located with the camera. Depending on the setup, the pupil is either very dark or very bright, and that is the signal used to extract the pupil from the rest of the image [10].

There are various challenges to pupil detection that can be studied, some of which are as follows:

- The view point may vary from sensor to sensor in pupil capturing.
- The person could be wearing glasses.
- The eyes could be partially closed.
- Sometime the eye lashes cover the eyes.
- There is also a lot of variations in the natural shape of the eyes.
- There can be issues of image quality, variation and lighting conditions.
- In some applications a lot of these variations can be controlled.

There are a number of pupil detection algorithms, such as classical algorithm edge detection, which can be followed by ellipse fitting to allow information from low level vision to be used to find the location of the pupil. The modern methods tend to be deep learning methods, for example NVGaze which uses CNN based methods **[11]**.

In addition, there are some artifacts which can be consider as presentation attack (PA) to spoof the liveness detection system. They are as follows:

### 2.1. Artifacts for GTPD based PAD

Gaze tracking is one of the challenges faced by response-based PAD for face recognition systems. In this method, researchers used an artifact mask or printed photo, as shown in Figure 2.

**Photo-printed attack:** attackers show facial images as a printed photo to spoof challenges in response-based recognition systems.

**Display attack:** attackers show facial images such as a displayed photo to spoof challenges in response-based recognition systems.

**Replay attack:** replaying a video and presenting it as a biometric is a replay attack. In this scenario, attackers use original video to generate fake eye tracking.

**2D mask attack:** attackers build 2D models of a real user, which the 2D model is used as presentation attack instrument (PAI).

**3D mask attack:** Attackers use silicon to make a 3D face of a real user in order to spoof biometric systems [12].

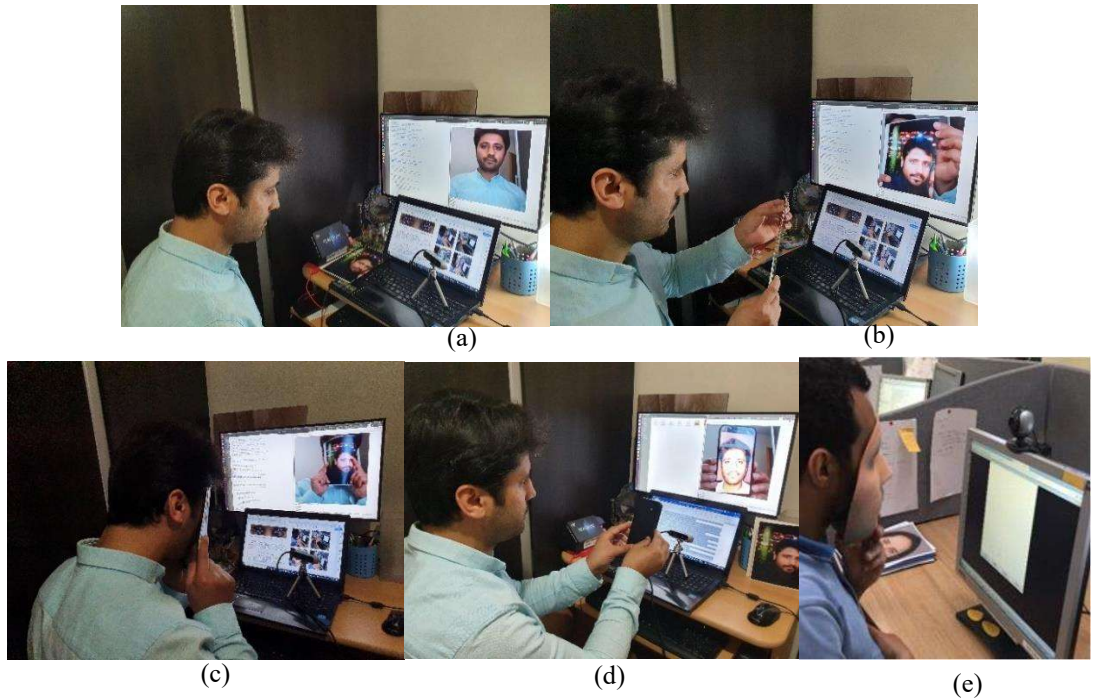

(a)　　　　　　　　　　　　　　　　(b)

(c)　　　　　　　　(d)　　　　　　　　(e)

**Fig. 2.** (a) Bona fide subject and instances of face PAs: (b) printed photo attack, (c) 2D mask attack, (d) display attack and (E) 3D mask attack (this photo is taken from Alsufyani et al. [13])

The first idea of pupil dynamics was patented by Andrzej Pacut et al. in 2007 [1]. They used pupil dynamics for iris liveness detection. Later on, in 2015, his colleague Adam Czaika explored pupil dynamics for iris liveness detection by introducing a database and a new evaluation metric [2].

**Contact lens:** Attackers manufacture an iris on a visual contact lens, as shown in Figure 3. The contact lens's texture overlays the real iris texture and a mixed texture is created. To the best of our knowledge, contact lenses have been used by attackers to spoof pupil dynamics systems, though no research has been published in this area.

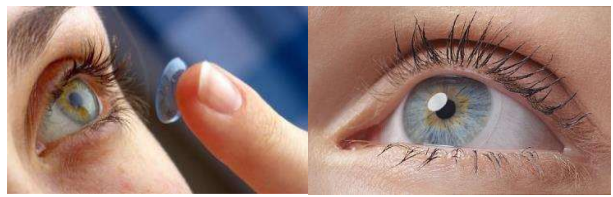

**Fig 3.** Contact lens on a human eye. (Image Source: Google)

## 3. PAD Methodology Preview

We use two different subsections to cover facial PAD and iris PAD. First we explain facial PAD using recent literature concerning eye movement, then we review iris PAD using eye movement and explore the relationship between these two topics.

### 3.1. PAD for facial recognition using eye movement

In [14] pupil movement is used to show whether an eye is real or artificial. Landmarks of the face and points around the eye are used to measure the distance between the eye and the pupil. Variations, minimum and maximum distances are extracted and then used to detecting PAs. This liveness detection method is used for protecting facial biometric systems. A 3D mask, 2D photo and displayed photo on a tablet are each used for PA. Eighty subjects participated in the data collection. The false negative rate (FNR) and false positive rate (FPR) were 10% and 8% respectively. The authors concluded that 3D mask attacks are more difficult to detect than other attacks.

Using iris movement for liveness detection is an idea to offer a secure PAD method. In [15] a method based on iris movement is proposed, called IriTrack. The eye should follow a randomly generated poly-line. In [15], three challenges were studied. First, existing noise in eye movements by user-device interaction; second, the difference in image transformation when images are captured; and third, the pattern has complex interaction that can reduce the number of successful attempts by attackers. They draw a pattern randomly on screen and the user should follow it. It works in a similar way to commonplace unlock pattern systems on mobile devices. However, in IriTrack, the pattern is followed by the eyes, then the cosine distance is used to measure similarity between the enrolled pattern and the presented pattern. They had 18 subjects with 40 presentations each, giving a dataset containing 720 real presentations. The time for performance on a PC with 16GB RAM and one Intel Dual-Core i7-6600U CPU takes 3.845 seconds. It is not possible to track eye angles on a small screen. The eye changes are very small, making it hard even for a large screen. If a pattern is generated randomly, it effectively prevents video attacks, because each attempt would be random.

Eye-glasses have an effect on PAD. For that reason, the authors in [16] proposed a method based on gaze information. 2D and 3D masks, as well as projected photos, were used for attack scenarios. The authors found that tinted glasses do not have a large impact on detection performance. In the proposed method, a visual stimulus appears on a display to capture data (facial images). To extract features – facial landmark points – Chehra Version 3.0 [17] is used, which extracts 59 landmarks. Moreover, to extract gaze-based features, the variance of the pupil center is measured. It is believed that there is variance between a genuine attempt and an attempt by an imposter. As a result, the variances are used to classify fake and genuine attempts. Eighty users participated in data collection. In addition, 2D mask, 3D mask and projected photo are all considered for PA. The results of the experiment showed that wearing glasses does not impact performance. The true positive rate (TPR) for the 2D mask, projected photo and 3D mask PA in 15 sets of collocated points with tinted glasses were 43%, 51% and 90%, respectively. Whereas the TPR for the attacks when the participants were not wearing tinted glasses were 43%, 57% and 93%, respectively.

Facial recognition was protected against attack presentation by utilizing pupil tracking in [18]. They used a Haar-Cascade Classifier [19] to detect the eye region and the Kanade-Lucas-Tomasi (KLT) [20] approach to track the eye. They then used the EyeMap method to extract the position of the pupil. To evaluate the proposed method, the authors used Yale Face Database part B and achieved a 98% success ratio.

### 3.2. PAD for iris recognition using eye movement

The authors in [1] believed that the existing liveness detection systems are not automatic nor accurate. They proposed a method based on pupil dynamics. In compliance with the method, the measurement of the characteristic dimensions of the hypothetical pupil is taken on the basis of a sequence of images. The eye is stimulated with light featuring a pre-defined intensity profile. For each image in this sequence, the characteristic dimensions of the hypothetical pupil are calculated by means of image processing methods. For a sequence of images, the system determines a function that defines the changes in the characteristic dimensions of the hypothetical pupil within the measurement period, and then, on the basis of the changes as well as on the selected mathematical model, the liveness parameters of the eye are determined using estimation methods. The calculated liveness parameters are compared with a statistical template by way of a classification process.

In [2] a method for eye liveness detection based on pupil dynamics was proposed. The author showed that the pupil reacts to visible light and its size can vary. Mimicking pupil dynamics for an artificial object is difficult. The author did not use a contact lens or paper printout for evaluation. Therefore, the author tried to classify spontaneous pupil oscillations and normal pupil reaction to a positive surge of visible light. For a better understanding of performance error, the author divided the dataset into two separate subsets, used to train and evaluate a given method. The method was performed on 26 independent subjects. This work presents research results for people who are not stressed and who have not ingested any substance that could modify pupil reaction.

In [3] Iris liveness detection based on gaze estimation is proposed. The authors collected eye movement and iris images from 100 subjects. By creating a hole in the place of the pupil on the printed images, they discovered that it was possible to spoof an iris and eye movement biometric system. To handle this problem, they used gaze estimation based on Pupil Center Corneal Reflection (PCCR) to develop a methodology for print attack detection. Their system comprised a camera and a light source. The BMT_20 Iris Recognition System was used to collect the real iris images and an HP LaserJet 4350dtn gray scale printer was used to collect iris print-attack images. In Spoofing Attack Scenario I (SAS-I), the employed eye tracking system was directly attacked using the prepared iris printouts, both during the calibration procedure and during the main gaze recording phase (stimulus presentation stage). Gaze signals that are captured during such an attack, should carry distortions that arise from both sources. In Spoofing Attack Scenario II (SAS-II), the employed eye tracking system was attacked exclusively during the stimulus presentation stage, whereas the calibration procedure was implemented by a valid (live) eye. This scenario corresponds to the case that the attacker is able to bypass the calibration procedure (e.g., the system is pre-calibrated) and performed the spoofing attack directly during the main process. For these types of spoofing attacks – SAS-I and SAS-II – an equal error rate (EER) of 6.6 and 5.9 respectively was reported. The system is slow and recognition details were not reported.

The authors in [21] presented a PAD method of visible iris recognition. They believed subtle phase information discriminates well between fake and real biometrics. They used Eulerian Video Magnification (EVM) to gather that information. The captured iris video is preprocessed to consist of 30 frames between blink intervals. The registered iris video is decomposed using the Fourier Transform to obtain the phase and magnitude component. The phase component of the iris video is magnified to emphasize the variation in the phase used to make the decision. EVM can magnify the temporal variations in video by decomposing it and applying a temporal filter. They believed that the replayed video presents different information and there will be a lot of noise. They used EVM to enhance small variations in the phase component of the video frame. The video frame was decomposed using Fourier Transform to obtain the phase information, which was then fed through EVM. The decomposed phase information was spatially filtered using Laplacian pyramids and temporally filtered using the Butterworth lowpass filter to magnify the variations in the phase of each frame in the video. The enhanced phase variation in the video was used to estimate the liveness of the subject as normal presentation. All the results related to the proposed scheme of PAD were disclosed in terms of the Attack Presentation Classification Error Rate (APCER) and the Normal Presentation Classification Error Rate (NPCER) [22]. APCER is defined as the proportion of attack presentations incorrectly classified as normal presentations in a specific scenario, while NPCER is defined as the proportion of normal presentations incorrectly classified as attack presentations in a specific scenario, [22]. Further, the authors also disclosed the results in terms of Average-Classification-Error-Rate (ACER), which is described as the average of APCER and NPCER. ACER is obtained on a testing database when different frames starting from 6 to 11 are considered with a threshold of Th = 0.7. The results were indicated from frame number 6, as the frames 1 to 5 were used to make the first decision at frame 6. From the obtained results, the best possible and reliable frame for making a decision was frame number 11, which provided an ACER of 0% for all cases. Table 1 illustrates existing methods for PAD using GTPD methods.

**Table 1**. An overview of algorithms for GTPD-based PAD methods

| Reference | Attacks | Techniques | Algorithm and methods | |
|-----------|---------|------------|----------------------|--|
| Asad Ali et al. [16] | Printed photo, 2D mask, 3D mask | Using a visual stimulus to track eyes. | Extract facial landmark | Chehra Version 3.0 [17] |
| | | | Pupil extraction | Propose a method based on variance, min and max between landmarks |
| Meng Shen et al. [15] | Printed photo, Replay video, 2D/3D mask | Iris movement | Attack detection | probability-based random pattern generation method |
| | | | Face detector | Haar classifiers |
| Killioglu et al. [18] | No attack | Eye area estimation | Eye area estimation | Haar-Cascade Classifier |
| | | | Eye-tracking | Kanade-Lucas-Tomasi (KLT) |
| | | | Pupils extraction | EyeMap method |
| Czajka et al. [1, 2, 23] | No attack | Using pupil dynamic | Pupil extraction | Kohn and Clynes [24] |
| | | | Live and fake classification | Support Vector Machine (SVM) |
| Rigas et al. [3] | Print-attacks | Gaze estimation | Gaze estimation | Pupil Center Corneal Reflection (PCCR) |
| | | | Liveness Detection | New version of the Complex Eye Movement (CEM) method |
| Raja et al. [21] | Artefact Video | video presentation attacks detection | PAD method | Modified Eulerian video magnification (EVM) |

## 4. Databases

To carry out research and develop a PAD method, it is necessary to have available databases. In this section, we present all the existing databases for liveness detection based on GTPD.

**Warsaw-BioBase-Pupil-Dynamics (V1.0, V2.0, and V3.0)**

This database is created for patenting pupil dynamic for iris liveness detection. The first version of the database was collected from 26 subjects with a 30-second timeline [2]. In each second, 25 frames were captured. This database was collected in the Warsaw University of Technology and Research and Academic Computer Network (NASK). To extend V1.0, 60 subjects were added in the second version, giving 86 subjects in total [23]. V1.0 and V2.0 are considered to be image-based databases. In the current version (V3.0), 42 videos from 42 subjects were recorded with a 30-second timeline [25]. All the versions were collected in the same scenario. To the best of our knowledge, it remains the best-known database for pupil dynamic liveness detection. The details are available at http://zbum.ia.pw.edu.pl/EN/node/46.

**Eye Tracker Print-Attack Database (ETPAD) v1**

Authors at Michigan State University created a database of 100 subjects who each participated in a data collection scenario twice [3]. They used EyeLink (EyeLink 1000 Eye Tracker), available at http://www.sr-research.com, to record live (genuine) data from the participants' eyes. The distance between the display and the participant's head was 0.55 m. To make a presentation attack, they used a printed-photo attack scenario, and to print images they used a HP Laserjet 4350dtn [3]. The details are available at https://userweb.cs.txstate.edu/~ok11/etpad_v1.html.

**Gaze Alignment Database**

Authors at the University of Kent built a database with 70 subjects from various countries for PAD based on gaze alignment. [14]. The database was collected to be used in face liveness detection. To make a presentation attack, they used four various attack presentation scenarios as follows:

- Display attack (iPad screen as a PAI)
- 2D printed photo attack (printing biometric on paper)
- 2D mask attack (making a 2D printed photo like a 3D mask)
- 3D mask attack (making an artificial face using silicon)

More details can be found in [14].

**IriTrack Database**

In this database, 18 volunteers participated 40 times in data collection scenarios for testing [15]. Overall, the database contains 720 recorded videos from genuine cases. It has been collected for use in face spoofing attacks. To make the presentation attacks, they used replay attack presentation. More details can be found in [15].

**Visible Spectrum Smartphone Iris Video (VSSIRISV) database**

Researchers at the Norwegian University of Science and Technology made a publicly available database for iris PAD based on mobile devices [21]. They collected biometrics from 31 participants. In total, there are 62 individual irises with a timeline of 1-4 seconds. To make presentation attacks, they replay videos to carry out replay attack presentations. The details are available at www.nislab.no/biometrics_lab/vssirisv_db.

Table 2 and Table 3 indicate brief details on existing databases for GTPD presentation attack detection methods that can be patterned in future researches. Table 2 shows databases from the same source, from which they made datasets for pupil dynamic iris PAD. The authors believe that making attack presentation for pupil dynamics is difficult and, to the best of our knowledge, there is no artifact data for pupil dynamic based databases. The field of biometric studies is very challenging in terms of data collection when genuine data is required. Even though collecting biometric data like face or image base biometric via the internet is easy, they cannot be used in presentation attack detection scenarios.

**Table 2**. An overview of existing databases for GTPD presentation attack detection without Attack presentations

| Dataset | # of Subjects | Recording time(s) | Sensor | #of samples | Distance sensor with user | Year and Ref. |
|---|---|---|---|---|---|---|
| **Warsaw BioBase Pupil Dynamics V1.0** | 26 | 30 | IrisCUBE camera, visible LED | 204 | 30cm | 2015,[1] |

| | | | | | | | |
|---|---|---|---|---|---|---|---|
| **Warsaw BioBase Pupil Dynamics V2.0** | 86 | 30 | IrisCUBE camera, visible LED | 435 | 30cm | 2019,[23] |
| **Warsaw BioBase Pupil Dynamics V3.0** | 42 | 30 | IrisCUBE camera, visible LED | 42 | 30cm | 2019,[25] |

**Table 3**. An overview of existing databases for GTPD presentation attack detection with Attack presentations

| Dataset | # of Subjects | Recording time(s) | Resolution (pixels) | Attacks | Sensor for fake data | # of samples | Sensor for genuine data | # of samples | Distance sensor with user | Year and Ref. |
|---|---|---|---|---|---|---|---|---|---|---|
| **Eye Tracker Print-Attack Database (ETPAD) v1** | 100 | 15 | 1680*1050 | Print-Attack | HP Laserjet 4350dtn | 200 | EyeLink 1000 | 200 | 55cm | 2014, [3] |
| **Gaze Alignment Database** | 70 | 10 | 640*480 | 2D mask 2D photo and 3D mask attack | iPad screen, a webcam | 210 | A webcam | 70 | - | 2018, [14] |
| **IriTrack Database** | 18 | 5 | 640*480 | Replay Attack | - | 720 | OV5693 sensor | 720 | 20 to36 cm | 2018, [15] |
| **Visible Spectrum Smartphone Iris Video (VSSIRISV) database** | 31 | 2-4 | - | Replay Attack | iPad (display). Nokia Lumia 1020 and iPhone 5S (recording device) | 124 | Nokia Lumia 1020 and iPhone 5S | 124 | 35 to 50 cm | 2015, [21] |

## 5.  Performance evaluation and metrics

According to the ISO/IEC DIS 30107-4 standard, PAD evaluation metrics are based on ISO/IEC 30107-1:2016 standard. Mobile devices are smartphones, laptop, wearable information devices, and tablet PCs. Matrices for mobile devices similar to PAD for non-portable devices are shown Table 4:

**Table 4.** *Matrices for mobile devices PAD*

| Metrices name | Acronym | Synonyms |
|---|---|---|
| Impostor Attack Presentation Accept Rate [2] | IAPAR | IAPAR shows percentage proportion of incorrectly classification of attack presentations as real presentation in liveness detection scenario. |
| Bona Fide Presentation Classification Error Rate [2] | BPCER | BPCER shows percentage of incorrectly classified of bona fide presentations as PA in liveness detection scenario |
| Attack Presentation Acquisition Rate [22] | APAR | APAR shows percentage of the data acquisition in biometric subsystem with considering a biometric instance of adequate quality when similar PAI species are used for attack presentations. |
| Impostor Attack Presentation Match Rate [22] | IAPMR | IAPMR shows percentage of matching target reference during impostor attack presentations. In this scenario similar PAI species is used in verification system |
| Concealer Attack Presentation Non-Match Rate [22] | CAPNMR | IAPMR shows percentage of the non-matching concealer reference during concealer attack presentations. In this scenario similar PAI species is used in verification system. |

There are some additional metrics that have been studied in recent eye-tracking-based research as follows:

- **Goodness of Fit (GoF):** a statistical model depicting how well a pupil liveness detection fits a set of pupil changes [2];
  - $GoF = \max(0, 1 - \frac{Norm_2(x - \hat{x})}{Norm_2(x - \bar{x})})$,
  - $x$: pupil size changes in real presentation
  - $\hat{x}$: model response
  - $\bar{x}$: mean of x
- **Half total error rates (HTER):** formed by summing the False genuine rate (FGR) and the False Fake Rate (FFR) and dividing it in two, as follows [13]:
  - $HTER = \frac{FGR + FFR}{2}$

- **Eye-corner Detection Accuracy (EcDA):** shows the proportion of genuine presentations correctly classified as correct following stimuli in eye-tracking scenarios [26].
- **False spoof acceptance rate (FSAR):** a method whereby imposter samples are classified as spoof samples [27]:
  - $FSAR = 100 * \frac{Imposter\ presentation\ classified\ as\ spoof\ samples}{All\ spoof\ samples}$
- **False live rejection rate (FLRR):** formed from the number of genuine samples indicated by presentation attack detection as a fake sample [27];
  - $FLRR = 100 * \frac{Imposter\ presentation\ classified\ as\ genuine\ samples}{All\ genuine\ samples}$
- **Classification rate (CR):** showing the number of test samples performance, where each sample was correctly classified as live or spoof. It is calculated as follows [27]:
  - $CR = 100 * \frac{Correctly\ Classified\ Samples}{All\ samples}$

### 5.1. Performance evaluation

Challenge based liveness detection methods are evaluated using several different metrics. We want to achieve new ideas in challenge-based liveness detection methods. In this work, we evaluate GTPD-based PAD algorithms. To achieve our goals, we assess six various state-of-the-art iris and face liveness detection methods. Various types of attacks were presented to spoof biometric systems, including printed photo attack, 2D mask attack, 3D mask attack and replay attack. Gaze tracking and pupil dynamics are two different methods for liveness detection. Therefore, the state-of-the-art methods utilized different type of attacks, databases, and liveness detection techniques. Consequently, we evaluate them separately. Pupil dynamics liveness detection methods are (a) eye area estimation and (b) measuring the pupil diameter. Gaze tracking methods are (a) tracking eyes on visual stimuli, (b) extracting iris movement, (c) eye area estimation, (d) video presentation attacks detection. Most of the research reviewed in this work used IAPAR and BPCER evaluation metrics, which are well-known metrics in the ISO/IEC standard.

Table 5 illustrates the performance of existing methods on various databases. In the work by Alsufyani et al., we observed that there are a number of line segmentations in the evaluation process. Hence, we just consider five lines and report the outcomes. In the study by Asad Ali et al., we found that they considered different timelines in the evaluation process. Hence, we take this into account and consider a 5-second timeline. In the paper by Rigas et al., they reported IAPAR and BPCER on different models, while we report here just iPhone against Nokia, and Nokia against iPhone. In the paper by Rigas et al., they have evaluated various scenarios and used different metrics which are the same with IAPAR and BPCER. Therefore, we just show Spoofing Attack Scenario I (SAS-I) results.

**Table 5**. The performance of existing methods on a number of databases

| Reference | Database name/Sensor | | Printed-photo attack | | 2D mask attack | | 3D mask attack | | Replay video attack | |
|---|---|---|---|---|---|---|---|---|---|---|
| | | | IAPAR | BPCER | IAPAR | BPCER | IAPAR | BPCER | IAPAR | BPCER |
| Alsufyani et al. [16] | Gaze Alignment Dataset | Tablet | 93% | 91% | 89% | 82% | 86% | 77% | - | - |
| | | Phone | 88% | 80% | 81% | 66% | 82% | 68% | | |
| Asad Ali et al. [14] | Gaze Alignment Dataset | Tablet | - | 80% | - | 98% | - | 98% | - | - |
| | | Phone | - | 74% | - | 99% | - | 99% | | |
| Rigas et al. [3] | Gaze Estimation Database | | 93.6% | 91.8% | - | - | - | - | - | - |
| Meng Shen et al. [15] | IriTrack Dataset | | - | - | - | - | - | - | 95.2% | 95.6% |
| Raja, et al. [21] | VSSIRISV | Nokia | - | - | - | - | - | - | 100% | 73.60% |
| | | iPhone | - | - | - | - | - | - | 100% | 73.62% |
| Adam Czajka [2] | Warsaw BioBase Pupil Dynamics V2.0 | | - | - | - | - | - | - | 100% | 100% |

To have a comprehensive view, we have done an evaluation on our dataset with the existing methods. The database has two part as follows:

A. Real data

In our data collection process, we have collected data from 56 participants. For each user we have asked them to participate in two different times (morning, evening). Before starting the data collection, we have explained every details and purpose of the data collection then we have started data collection. Every user has sat in a comfortable chair and fixed position. It means the head of the users have been fixed. In the first session data have been collected in two different conditions, first without light and second with light. Light have been switched on, when participants looking at the center. This is done just to see pupil's changes when there is light and there is no light. In the evening session, data collection process has been done just once. We have asked all participant to follow the same stimuli in both sessions. The order of the shapes can be completely random. We used an order of shapes on the screen (one for morning session, the other for evening session) for all users to see variety of eye movement and pupil diameter in different participant, when the condition is equal for all participants. Totally we have recorded 225 videos which every video is ~15 seconds. We used Samsung Galaxy S10+ to record videos from participants.

B. Fake data

To collect fake samples as imposter data, we used a new brand smartphone (Samsung Galaxy S10+), and the PC's display monitor. The captured videos for real samples have been shown on the display and are recorded with the Samsung Galaxy S10+. Totally, we have made 225 fake samples.

As it is mentioned earlier there is no database for working on gaze information and pupil dynamic parallelly. Therefore, our database provides this opportunity to evaluate the existing method. We have executed the existing methods on our database and the result has been depicted on Table 6.

**Table 6**: The HTER values for different methods on our database in different time sequence.

| Method name | HTER (%) | | |
|---|---|---|---|
| | 5 second | 10 second | 15 second |
| The angular difference extraction [11] | 19.4 | 17.5 | 16.02 |
| IriTrack [13] | 18.2 | 15.3 | 13.05 |
| Magnified Phase Information [20] | 20.4 | **14.3** | 12.5 |
| Pupil Center Corneal Reflection [3] | **17.03** | 15.4 | **12.02** |

| Pupil Dynamics [2] | 20.6 | 16.7 | 13.01 |
|---|---|---|---|

We reported HTER rate for all methods on our database. First column illustrates existing methods, all names are extracted from related paper, and they have been assessed on different database as we mentioned earlier. The second column depicts HTER in different time sequences. By using just first 5 second and whole part (15 seconds) of videos, Pupil Center Corneal Reflection method has achieved the best results with HTER=17.03 and HTER=12.02. However, Magnified Phase Information method with HTER=14.3 was the best method for having 10 second videos per samples. Some of methods are proposed for NIR images and had problem when the flash light was turned off. The database is recorded by smartphone camera which is visible spectrum. As result, it is hard to recognize and distinguish between pupil and iris. For instance, Pupil dynamic along with IriTrack method had problem when users were looking left or right because extracting pupil size and iris itself were difficult and had lots of noise.

## 6. Future research

The topic of challenge-response based presentation attack detection for facial and iris biometrics is well-known and seeks to find the most secure method. However, there are certain disadvantages in the way, such as user inconvenience and the cost computational problem. The topic still needs to be explored and investigated more deeply. On the other hand, it has been shown that challenge-based PAD can be used in real applications. For instance, Fujitsu has started to use eye blink technology to unlock smartphones based on iris recognition technology [28]. Here we present open issues and challenges to iris and facial PAD based on GTPD.

**Lack of generalization for different attacks**

Each biometric system has an artifact to spoof the system. For instance, to spoof facial recognition we use printed-photo, 2D mask, 3D mask, morphing, make up and etc. Moreover, to spoof iris recognition we use contact lenses, printouts of the iris and so on. GTPD, like other PAD methods, are fighting against presentation attacks. To spoof gaze tracking methods, attackers usually utilize replay videos. Meanwhile, there are also other potential possibilities, including the use of makeup and deep fake technology to spoof gaze tracking, but these have not been studied. Considering the pupil dynamics method, attackers can potentially spoof liveness detection based on pupil dynamics by using special contact lenses, though this has not been the subject of any studies yet.

**Databases and evaluation metrics**

It has been proved that there is a body of research into GTPD-based PAD, though the studies did not use the same evaluation metrics. In addition, there is an evaluation metrics standard for PAD called ISO/IEC 30107-3 [22] and ISO/IEC DIS 30107-4 [29]. This can throw up complications when looking at an overview of the research, for example, Farzin Deravi [4, 13, 14, 16] has more than four published papers in this field, but there is no similarity between his evaluation metrics and the standard metrics. Hence, we need to make a comprehensive study for GTPD-based PAD. In terms of database availability, we do not have a complete available database in both gaze tracking and pupil dynamic base liveness detection.

**Independence between iris and facial recognition system and PAD**

Presentation attack detection is a separate topic to recognition. This means, we should study these topics separately, because liveness detection is at a sensor level, in contrast to the recognition systems. There is an idea to merge PAD and recognition systems, for instance obtaining input features and recognizing the input as fake, genuine or imposter at the same level of processing.

**7. Conclusions**

In conclusion, we gathered all existing GTPD methods and indicated that having a challenge-response based method for liveness detection in biometric recognition systems could decrease the vulnerability of the system against presentation attacks. However, there were lots of challenges and issues that should be addressed properly. For instance, the lack of generalization for different methods of attacks, the lack of complete databases, and so on. We discussed all kinds of attack for GTPD, and looked at how to prevent against these attacks. We reviewed all the available databases for gaze tracking and pupil dynamics. An interesting aspect of this work was about pupil dynamics, an area that has not been studied a lot. Therefore, we had a huge gap to fill in the pupil dynamics liveness detection field of study. We provided an overview of performance evaluation metrics introduced in ISO/IEC 30107-3 and ISO/IEC DIS 30107-4. However, there were more useful metrics that we discovered during the course of this work. Finally, it might be observed that liveness detection and recognition system are independent topics. We believed that they could probably be merged together and, based on this idea, we could decrease the computational cost of the system. GTPD liveness detection has a large number of potential applications, in particular controlling security in industries that are high risk places. In addition, it is trustable and can be used in ATMs etc.

## CONFLICTS OF INTEREST

The corresponding author states that there is no conflict of interest.

## ACKNOWLEDGMENT

This paper was supported by the European Union's Horizon 2020 research and innovation programmed under the Marie Skłodowska-Curie grant agreement No 675087.

# Author biography

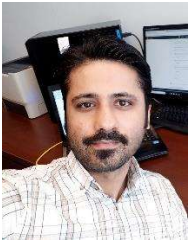 **JALIL NOURMOHAMMADI KHIARAK** was born in Ardabil, Iran in 1989. He received the B.Sc. degree from the non-Profit University of Sabalan, Ardabil, Iran, in 2011, and the M.Sc. degree in Artificial Intelligence from the Faculty of Electrical & Computer Engineering, University of Tabriz, Tabriz, Iran in 2015. His Master's thesis was entitled "Learning Hierarchical Representations for Video Analysis Using Deep Learning". His research interests are Biometric, Machine Learning, Computer Vision, Supervised learning (SVM, Back propagation, Logistic regression), unsupervised learning (Deep learning such as Auto-Encoders and Convolutional neural networks, PCA, K-mean Clustering), and Neural Networks. In December 2017 Jalil joined the Amber project at the Warsaw university of Technology in which he will be working on "Countermeasure algorithms against subterfuge in mobile biometric systems". He is currently member of European Association for Biometrics and Marie Curie Alumni Association.

**CONFLICTS OF INTEREST**

The corresponding author states that there is no conflict of interest.